\documentclass[journal,twoside,web]{ieeecolor}
\usepackage{generic}
\usepackage{cite}
\usepackage{amsmath,amssymb,amsfonts}
\usepackage{algorithmic}
\usepackage{graphicx}
\usepackage{algorithm,algorithmic}
\usepackage{hyperref}
\hypersetup{hidelinks=true}
\usepackage{textcomp}
\usepackage{fancyhdr}
\usepackage{epstopdf}
\usepackage{booktabs}
\usepackage{tabularx}
\usepackage{subcaption}
\usepackage{multirow}
\usepackage{array}

\let\labelindent\relax
\usepackage{enumitem}
\def\BibTeX{{\rm B\kern-.05em{\sc i\kern-.025em b}\kern-.08em
    T\kern-.1667em\lower.7ex\hbox{E}\kern-.125emX}}
\begin{document}

\title{Counterfactual Explainability Framework With CycleGAN And Counterfactual-Classifier Alignment Score For Retinal Disease Detection}

\author{Kritanu Chattopadhyay, Sayanjit Singha Roy, \IEEEmembership{Member, IEEE}, and Soumya Chatterjee, \IEEEmembership{Member, IEEE}\vspace{-2.2em}%
\thanks{Kritanu Chattopadhyay is with the Department of Mechanical Engineering,
National Institute of Technology Durgapur, West Bengal, 713209, India. 
(e-mail: kritanuchattopadhyaywork@gmail.com).}
\thanks{Sayanjit Singha Roy (\textit{Corresponding author}) is with the Department of Electrical Engineering, IIT Kanpur, Uttar Pradesh, 208016, India. 
(e-mail: sayanjit2011@gmail.com).}
\thanks{Soumya Chatterjee is with the Department of Electrical Engineering,
National Institute of Technology Durgapur, West Bengal, 713209, India. 
(e-mail: chapeshwar@gmail.com).}}

\maketitle
\pagestyle{fancy}
\fancyhf{}
\fancyhead[L]{\footnotesize \thepage}
\fancyhead[R]{\footnotesize IEEE Journal and Transactions Template,~Vol.~XX, No.~XX, 2026}
\renewcommand{\headrulewidth}{0.4pt}
\renewcommand{\footrulewidth}{0pt}
\setlength{\headheight}{14pt}

\begin{abstract}
Automated detection of vision impairing retina-based ocular conditions from fundus images is important for early screening, timely referral and reducing dependency on specialist-only assessment, for which neural network-based deep learning (DL) models have been widely utilized. However, explainability of the DL frameworks remains a major bottleneck for clinical adoption, particularly when model decisions are not linked to retinal regions that are clinically meaningful. To address this issue, this study presents CounterFundus, a novel CycleGAN-driven counterfactual explainability framework, integrating EfficientNet-B5-based retinal disease detection with visually interpretable disease-to-normal fundus image translation. For each pathological image, the counterfactual yielded by the CycleGAN generator represents an estimated healthy counterpart and the resultant difference map is utilized to localize disease-associated retinal changes. Unlike conventional post-hoc saliency methods, CounterFundus provides counterfactual explanations through visually plausible disease-to-normal retinal translation. Thereafter, to quantify the spatial agreement between counterfactual difference maps and classifier saliency, the Counterfactual-Classifier Alignment Score (CCAS) is introduced, embedding Spearman correlation, binary IoU and pointing accuracy into a single assessment protocol. To this end, EigenCAM-aligned evaluation demonstrates that the generated counterfactual explanations remain spatially consistent with classifier-relevant retinal evidence across all CCAS dimensions. Along with that, ablation studies further confirm that CCAS-filtered counterfactual augmentation improves the downstream classification performance in fundus images, establishing CounterFundus as a clinically-grounded, explainable artificially intelligence (XAI) framework for retinal disease detection.
\end{abstract}


\section{Introduction}
\label{sec:introduction}
\IEEEPARstart{F}{undus} imaging serves as a primary diagnostic tool for a range of vision-threatening retinal conditions including diabetic retinopathy, glaucoma and cataract. For the purpose of automated retinal disease detection, neural network (NN)-based deep learning (DL) frameworks have demonstrated strong quantitative performance, yet clinical translation remains severely limited by the explainability of the DL models due to their black box-like decision making mechanism \cite{liao2023attention, patrizi2022virtual}. Also, clinicians require spatially grounded evidence behind pathological predictions, whereas most post-hoc explanation methods fall short, providing limited clinical interpretability. \cite{jiang2025cnn}. In this context, gradient-based attribution methods such as GradCAM \cite{sharma2025improved} and its variants \cite{guo2023analysis} can highlight discriminative regions within the classifiers' feature space but are fundamentally tied to the model's internal gradients and offer no insight how the same fundus image would appear without pathological changes. Perturbation-based methods, such as LIME \cite{cao2025comprehensive}, rely on superpixel masking, where the segmented regions are often biologically arbitrary and may not correspond to clinically meaningful structures such as the optic disc, macula or vessel arcades. Therefore, the research gap between saliency maps and clinical semantics remains unaddressed, particularly in multi-class retinal disease detection.

To this end, counterfactual explanations offer a fundamentally novel paradigm to model interpretation by identifying the changes in an input that would alter the model decision \cite{wachter2018counterfactualexplanationsopeningblack}. In fundus imaging, this can be formulated as generating a visually plausible healthy counterpart of a diseased retinal image along with utilizing the pixel-level difference to localize disease-related regions. However, existing counterfactual studies do not quantify how well these difference maps align with the classifier-yielded disease-relevant regions.

\section{Background and Scope of Work}
Counterfactual explanations were formalised by Wachter $et$ $al$. \cite{wachter2018counterfactualexplanationsopeningblack} as minimal-change inputs capable of altering a model's decision, establishing the interpretable paradigm on which CounterFundus is built. Since then, the literature has mainly evolved along three directions; Assessing saliency reliability, generating realistic counterfactual images using generative models and validating whether the explanations are clinically meaningful. The first direction focuses on assessing saliency reliability. Through IoU-based analysis, Arun $et$ $al$. \cite{arun2021assessinguntrustworthinesssaliencymaps} showed that saliency methods have limited localization efficiency in medical imaging, directly motivating the spatial alignment protocol proposed in this work.

Focusing on generating realistic counterfactual images through generative models, GAN-based counterfactual generation was later explored by Mertes $et$ $al$. \cite{Mertes2022GANterfactual}, while Narayanaswamy $et$ $al$. \cite{narayanaswamy2023counterfactual} extended this direction using CycleGAN-based abnormal-to-normal image translation. More recently, diffusion models have emerged as an alternative to GAN-based counterfactual generation, with Atad $et$ $al$. \cite{Atad_2024} implementing a diffusion autoencoder for RetinaMNIST and Zigutyte $et$ $al$. \cite{Zigutyte2025CounterfactualDiffusion} using DDIM conditioning for histopathology explainability. In parallel, CycleGAN \cite{zhu2020unpairedimagetoimagetranslationusing} has also shown substantial efficacy in fundus domain adaptation \cite{yang2020residualcycleganbasedcameraadaptation} as well as in multi-modal retinal image synthesis \cite{Sindel_2023}. Moreover, Xue $et$ $al$. \cite{xue2019selectivesyntheticaugmentationquality} further introduced quality-driven synthetic augmentation, which is conceptually related to the CCAS-based filtering strategy implemented in the present work.

From the clinical validation perspective, prior studies have increasingly emphasized that explanation maps should not only be visually plausible but also correspond to disease-relevant lesion-level evidence. Calibration-based quality control for chest X-ray counterfactuals was introduced in \cite{Thiagarajan2022CalibrationCounterfactual}, highlighting the need for joint evaluation of visual interpretability and explanation reliability. In retinal disease detection, Boreiko $et$ $al$. \cite{Boreiko2022VisualExplanations} generated visual counterfactual explanations for diabetic retinopathy and evaluated them against expert lesion masks using IoU, making it one of the closest prior studies to the present work. Djoumessi $et$ $al$. \cite{Djoumessi2024.06.27.24309574} further demonstrated that interpretable diabetic retinopathy models with lesion-level evidence maps can improve screening speed and diagnostic accuracy, supporting the need for XAI frameworks that are visually feasible and spatially meaningful from a clinical perspective. EfficientNet-based backbones \cite{zhao2023transfsm} are also been widely used owing to their strong accuracy-to-complexity balance. Bajwa $et$ $al$. \cite{Bajwa_2019} addressed glaucoma localization, while 95.12\% accuracy has been reported in \cite{jimaging11080279} on the same 4-class Kaggle fundus dataset used in this study.

\begin{table}[!t]
\vspace{-0.8em}
\centering
\caption{Augmentation Strategy}
\label{tab:augmentation}
\setlength{\tabcolsep}{4pt}
\renewcommand{\arraystretch}{1.05}
\small
\begin{tabularx}{\columnwidth}{lXX}
\toprule
\textbf{Transform} & \textbf{Classifier Training} & \textbf{CycleGAN Training} \\
\midrule
Resize          & $224 \times 224$ & $224 \times 224$ \\
Horizontal Flip & $p=0.5$ & $p=0.5$ \\
Vertical Flip   & $p=0.3$ & $p=0.5$ \\
Rotation        & $15^\circ$ & $10^\circ$ \\
Color Jitter    & brightness = 0.2, contrast = 0.2 & brightness = 0.1, contrast = 0.1 \\
Normalization   & ImageNet mean/std & $[-1,1]$ \\
\bottomrule
\end{tabularx}
\vspace{-1.5em}
\end{table}

Despite these advances, a research gap still remains in evaluating counterfactual explanations in terms of both visual interpretability as well as spatial agreement with classifier-relevant retinal regions. To address this gap, the present study proposes CounterFundus, a CycleGAN-driven counterfactual XAI framework that integrates disease-to-normal retinal translation with classifier-aligned explanation assessment. Key highlights of the current work are mentioned below.

\begin{enumerate}[leftmargin=*, label=\arabic*), itemsep=0.2em, topsep=0.2em]
\item A CycleGAN-driven disease-to-normal retinal image translation framework is incorporated to generate visually plausible healthy counterparts of pathological fundus images.

\item A novel CCAS is introduced by embedding Spearman correlation, binary IoU and pointing accuracy into a single evaluation protocol for assessing the spatial agreement between counterfactual difference maps and classifier-relevant retinal regions.

\item A CCAS-filtered counterfactual augmentation strategy is implemented to retain spatially reliable synthetic samples for EfficientNet-B5-based retinal disease classification.

\item EigenCAM-guided classifier saliency is incorporated within the CCAS protocol to evaluate the alignment between generated counterfactual explanations and decision-relevant retinal evidence.

\item The proposed CounterFundus framework is designed as an XAI approach to generate spatially significant and clinically interpretable explanations for multi-class retinal disease detection. 
\end{enumerate}

\begin{figure}[!t]
\centering
\includegraphics[
width=0.7\columnwidth,
height=0.45\columnwidth,
keepaspectratio=false,
]{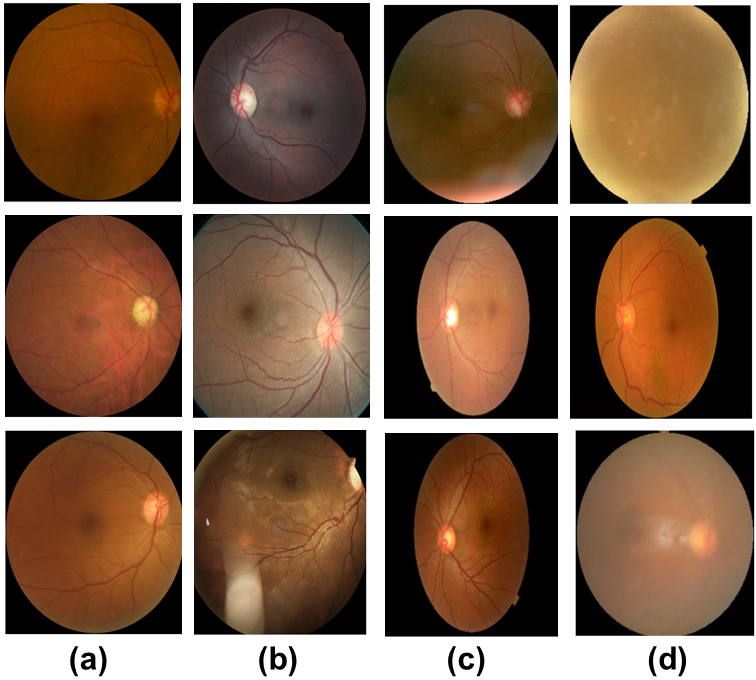}
\caption{Fundus images of (a) normal, (b) diabetic retinopathy, (c) glaucoma and (d) cataract categories, respectively.}
\label{fig:sample_images}
\vspace{-1em}
\end{figure}

\vspace{-0.5em}
\section{Dataset and Pre-processing}
\subsection{Fundus Image Dataset}
In this work, an online benchmark fundus image dataset \cite{DoddiEyeDiseasesDataset} is utilized, comprising 4217 RGB fundus images from four categories $viz.$ normal, diabetic retinopathy, glaucoma and cataract. The corresponding class-wise image counts are 1074, 1098, 1007 and 1038, respectively, indicating a mildly imbalanced distribution. The fundus images were captured under standard clinical imaging protocols at varying resolutions. Representative samples from the different fundus image categories are shown in Fig.~\ref{fig:sample_images}. To preserve the natural data distribution during evaluation, the mild class imbalance is handled through stratified sampling rather than oversampling.

\vspace{-0.5em}
\subsection{Image Augmentation and Cross-validation Splits}
The acquired fundus image dataset is first divided using a stratified holdout strategy, where 20\% of the images are reserved as a locked test set before model training. The remaining 80\% of the dataset is used for 5-fold stratified cross-validation (CV). During training, image augmentation is applied only to the training folds to synthetically increase the number of input images using the augmentation strategies listed in Table I. The validation and test sets are only resized and normalized, with no augmentation applied.

\vspace{-0.5em}
\section{Methodology}

\begin{figure*}[!t]
\vspace{-0.5em}
\centering
\includegraphics[
width=1\textwidth,
height=0.24\textheight,
keepaspectratio=false]
{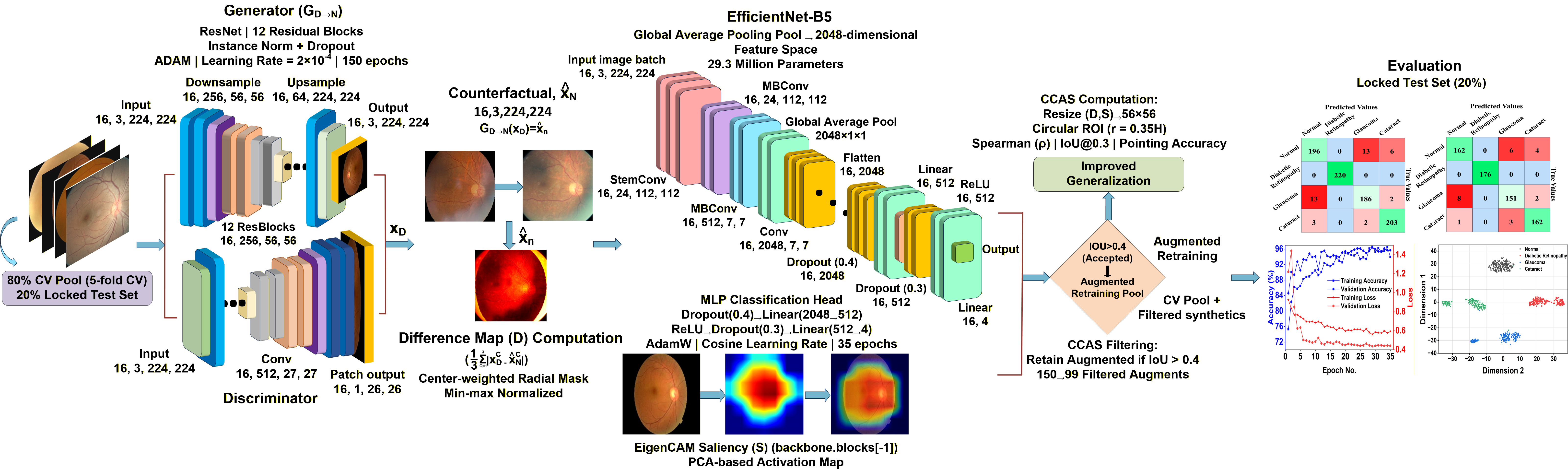}
\caption{Proposed CounterFundus pipeline.}
\label{fig:pipeline}
\vspace{-1em}
\end{figure*}

\subsection{Classification Backbone}
The proposed CounterFundus framework implements an EfficientNet-B5-based classification backbone \cite{zhao2023transfsm}, where the original classification head is replaced by a two-layer multi-layer perceptron (MLP). As can be seen from the CounterFundus workflow in Fig.~\ref{fig:pipeline}, the backbone extracts a 2048-dimensional feature, which is passed through a dropout-linear-ReLU-dropout-linear classification head with forward pass:
\begin{equation}
\hat{y} = W_2 \cdot \text{Dropout}\left(\text{ReLU}\left(W_1 \cdot \text{Dropout}(f_\theta(x))\right)\right)
\end{equation}

where, \(x \in \mathbb{R}^{3 \times 224 \times 224}\) denotes the input fundus image and $f_{\theta}(x) \in \mathbb{R}^{2048}$ represents the global average-pooled feature vector extracted by the EfficientNet-B5 backbone, parameterized by $\theta$. $W_1 \in \mathbb{R}^{512 \times 2048}$ projects the 2048-dimensional feature vector into a 512-dimensional latent representation, while $W_2 \in \mathbb{R}^{4 \times 512}$ maps the transformed features to the final disease-class output $\hat{y}$. Dropout is applied before the first and final linear layers with probabilities 0.4 and 0.3, respectively. ReLU introduces non-linearity between the two linear layers, while dropout reduces overfitting during training. Overall, the model contains approximately 29.3 million parameters.

\vspace{-0.5em}
\subsection{CycleGAN-based Counterfactual Generator}
The CycleGAN module \cite{zhu2020unpairedimagetoimagetranslationusing} learns bi-directional translation between diseased and normal fundus image domains. It consists of two generators, $G_{D \to N}$ and $G_{N \to D}$, and two discriminators, $D_N$ and $D_D$, for normal and diseased domains, respectively. Each generator follows a ResNet-based architecture with reflection padding, two downsampling layers, 12 residual blocks with instance normalization and dropout and two final upsampling layers followed by $Tanh$ activation. The discriminators use a 5-layer PatchGAN architecture with LeakyReLU activations. An image replay buffer of size 50 is also used improve training stability.

The generator objective combines adversarial, cycle-consistency and identity losses, expressed as:
\begin{equation}
\small
\mathcal{L}_G =
\mathcal{L}_{adv}
+ \lambda_{cyc}\left(\mathcal{L}_{cyc}^{D}
+ \mathcal{L}_{cyc}^{N}\right)
+ \lambda_{cyc}\lambda_{idt}
\left(\mathcal{L}_{idt}^{N}
+ \mathcal{L}_{idt}^{D}\right)
\end{equation}

where, $\lambda_{cyc}(=7)$ and $\lambda_{idt}(=0.3)$ control the contributions of cycle-consistency and identity preservation, respectively. The adversarial loss is measured using an MSE-based LSGAN objective, while the cycle-consistency term is computed using L1 loss. The network operates on $224\times224\times3$-dimensional fundus images using a base filter size of 64 and ADAM optimizer with $\beta_1 (= 0.5)$ and $\beta_2 (= 0.999)$. Model training is performed for 150 epochs with a learning rate of $2 \times 10^{-4}$ and a linear learning-rate decay is applied after epoch 75.

\vspace{-0.5em}
\subsection{Difference Map Computation}
For a diseased fundus image $x_D$, the corresponding normal counterfactual is generated as $\hat{x_N}=G_{D \to N}(x_D)$. The disease-associated changes are then localized by computing the channel-averaged absolute difference between the original diseased image and its generated counterfactual, expressed as:
\begin{equation}
\mathcal{D}_{raw}(i,j) = \frac{1}{3}\sum_{c=1}^{3} |x_D^c(i,j) - \hat{x}_N^c(i,j)|
\end{equation}

where, $i$ and $j$ denote the spatial pixel locations and $c$ represents the RGB channel index. To reduce peripheral boundary artifacts, the raw difference map $D_{raw}$ is weighted using a center-focused radial mask:
\begin{equation}
w(i,j) = 1 - 0.5 \cdot \frac{d(i,j)}{d_{max}}
\end{equation}

where, $d(i,j)$ is the Euclidean distance of pixel $(i,j)$ from the image center and $d_{max}$ denotes the maximum radial distance within the image. The final counterfactual difference incorporates radial weighting followed by min-max normalization:
\begin{equation}
\mathcal{D}(i,j) =
\frac{\mathcal{D}_{raw}(i,j) \cdot w(i,j) - \min}
{\max - \min + \epsilon}
\end{equation}

where, $\epsilon (= 10^{-8})$ is used for numerical stability. This weight-normalization step particularly suppresses peripheral noise while preserving clinically relevant retinal regions in the fundus images, including the optic disc and macula.

\vspace{-0.5em}
\subsection{Counterfactual-Classifier Alignment Score (CCAS)}
To quantify interpretability, the proposed CCAS measures spatial agreement between the counterfactual difference map, $D$ and the EigenCAM saliency map, $S$. Before computation, both maps are resized to $56 \times 56$ and evaluated within a central circular region corresponding to the retinal field of view. The CCAS protocol combines three metrics. First, Spearman rank correlation ($\rho$) evaluates the monotonic agreement between the intensity distributions of $D$ and $S$, defined as:
\begin{equation}
\rho = 1 - \frac{6\sum d_i^2}{n(n^2 - 1)}
\end{equation}

where, $d_i$ denotes the rank difference between corresponding pixels and $n$ is the number of evaluated pixels. Second, binary IoU measures the spatial overlap between thresholded salient regions, expressed as:
\begin{equation}
\text{IoU} =
\frac{\sum \mathbf{1}[\mathcal{D} > \tau] \cdot \mathbf{1}[\mathcal{S} > \tau]}
{\sum \mathbf{1}[(\mathcal{D} > \tau) \cup (\mathcal{S} > \tau)] + \epsilon}
\end{equation}

where, $\tau$ is the saliency threshold and $\epsilon$ prevents numerical instability. Finally, pointing accuracy ($PA$) evaluates whether the peak response locations of the two maps are spatially close:
\begin{equation}
\text{PA} = \mathbf{1}\left[||p_\mathcal{D} - p_\mathcal{S}||_2 < 0.2 \times H\right]
\end{equation}

where, $p_\mathcal{D}$ and $p_\mathcal{S}$ denote the peak locations in the counterfactual difference and EigenCAM maps, respectively and $H$ is the map height. The final CCAS is summarized across disease classes using 95\% confidence intervals.

\begin{table}[!t]
\centering
\caption{Fold-wise CounterFundus Performance}
\label{tab:clf}
\setlength{\tabcolsep}{4pt}
\renewcommand{\arraystretch}{1.08}
\small
\begin{tabular}{lccc}
\toprule
\textbf{Split} & \textbf{Acc. (\%)} & \textbf{AUC (\%)} & \textbf{F1 (\%)} \\
\midrule
Fold 1 & 94.52 & 99.04 & 94.44 \\
Fold 2 & 96.44 & 99.19 & 96.39 \\
Fold 3 & 95.41 & 99.05 & 95.34 \\
Fold 4 & 94.96 & 99.39 & 94.92 \\
Fold 5 & 94.66 & 99.19 & 94.59 \\
\midrule
CV & $95.2 \pm 0.69$ & $99.17 \pm 0.13$ & $95.14 \pm 0.7$ \\
\textbf{Test} & \textbf{95.38} & \textbf{99.69} & \textbf{95.31} \\
\bottomrule
\end{tabular}
\vspace{-1.5em}
\end{table}

\vspace{-0.5em}
\subsection{Training Protocol}
Training is performed using the AdamW optimizer with a learning rate of $1 \times 10^{-4}$, weight decay of $1 \times 10^{-4}$ and a batch size of 16. A cosine annealing schedule is applied over 35 epochs with $T_{\max}(=35)$. To reduce overconfident predictions, cross-entropy loss with label smoothing $\epsilon(=0.1)$ is used. In addition, mixup augmentation is employed with $\alpha(=0.4)$ to improve generalization, where the mixing coefficient $\lambda$ is sampled from a $\beta$-distribution and adjusted to retain one dominant class. The corresponding mixup loss is defined as:
\begin{equation}
\mathcal{L}_{mix} = \lambda \mathcal{L}(f(x_i), y_i) + (1-\lambda)\mathcal{L}(f(x_i), y_j)
\end{equation}

where, $x_i$ denotes the input image, $y_i$ and $y_j$ are the paired class labels and $f(\cdot)$ represents the classifier output. Gradient clipping with $\mathrm{max_{norm}}(=1)$ is further applied to improve training stability. Finally, 5-fold stratified CV is performed on the training set and the best model is selected based on validation accuracy.

\section{Results and Discussion}

\subsection{Disease Detection Performance of the Proposed CounterFundus Framework}

\begin{figure}[!t]
\centering
\setlength{\abovecaptionskip}{2pt}
\setlength{\belowcaptionskip}{2pt}
\captionsetup[subfigure]{labelformat=parens,labelsep=none,font=footnotesize}

\begin{subfigure}[t]{0.45\columnwidth}
\centering
\includegraphics[
width=\linewidth,
height=0.85\linewidth,
keepaspectratio=false,
clip
]{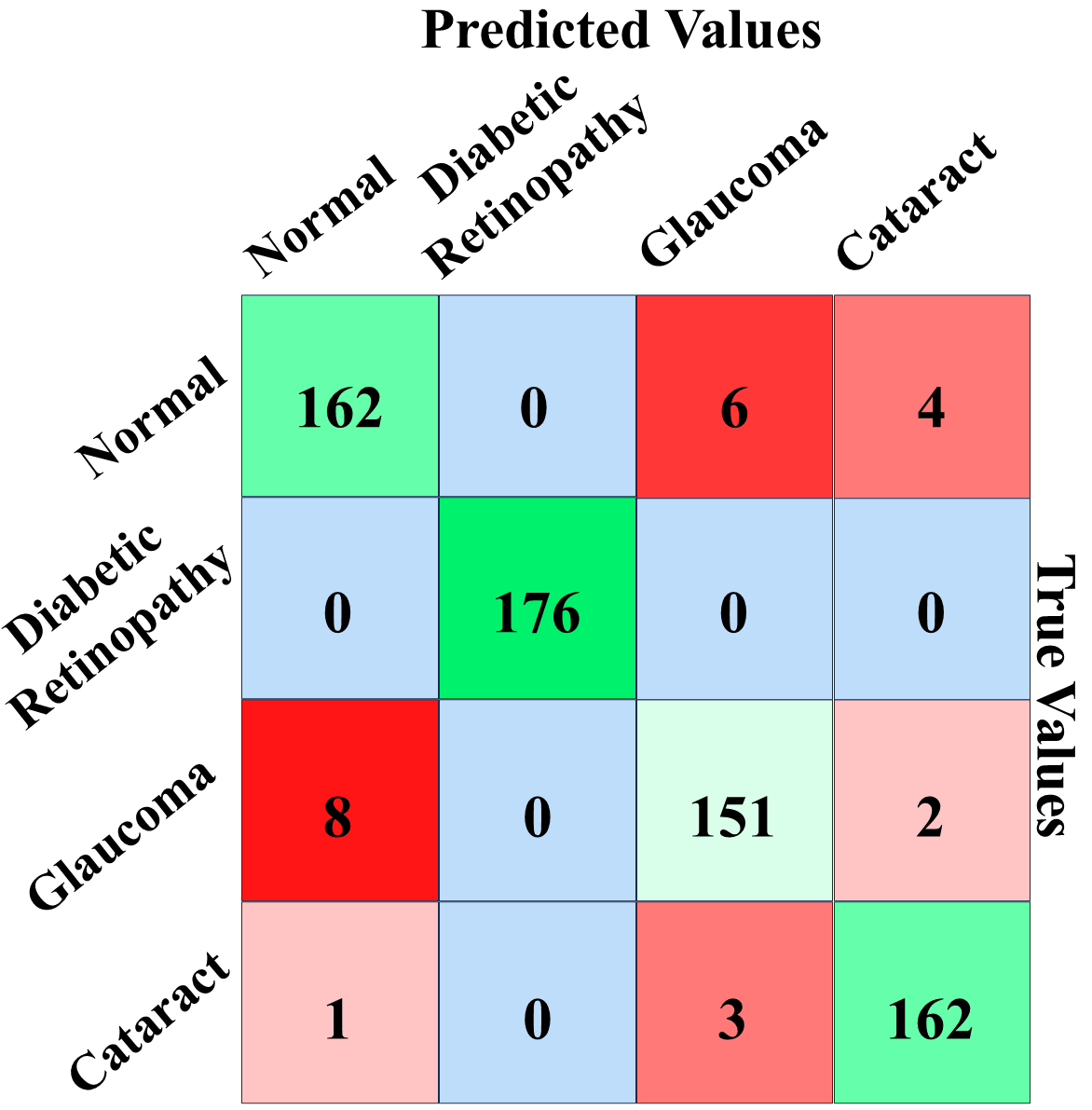}
\caption{}
\label{fig:best_fold_cm}
\end{subfigure}
\hfill
\begin{subfigure}[t]{0.45\columnwidth}
\centering
\includegraphics[
width=\linewidth,
height=0.85\linewidth,
keepaspectratio=false,
clip
]{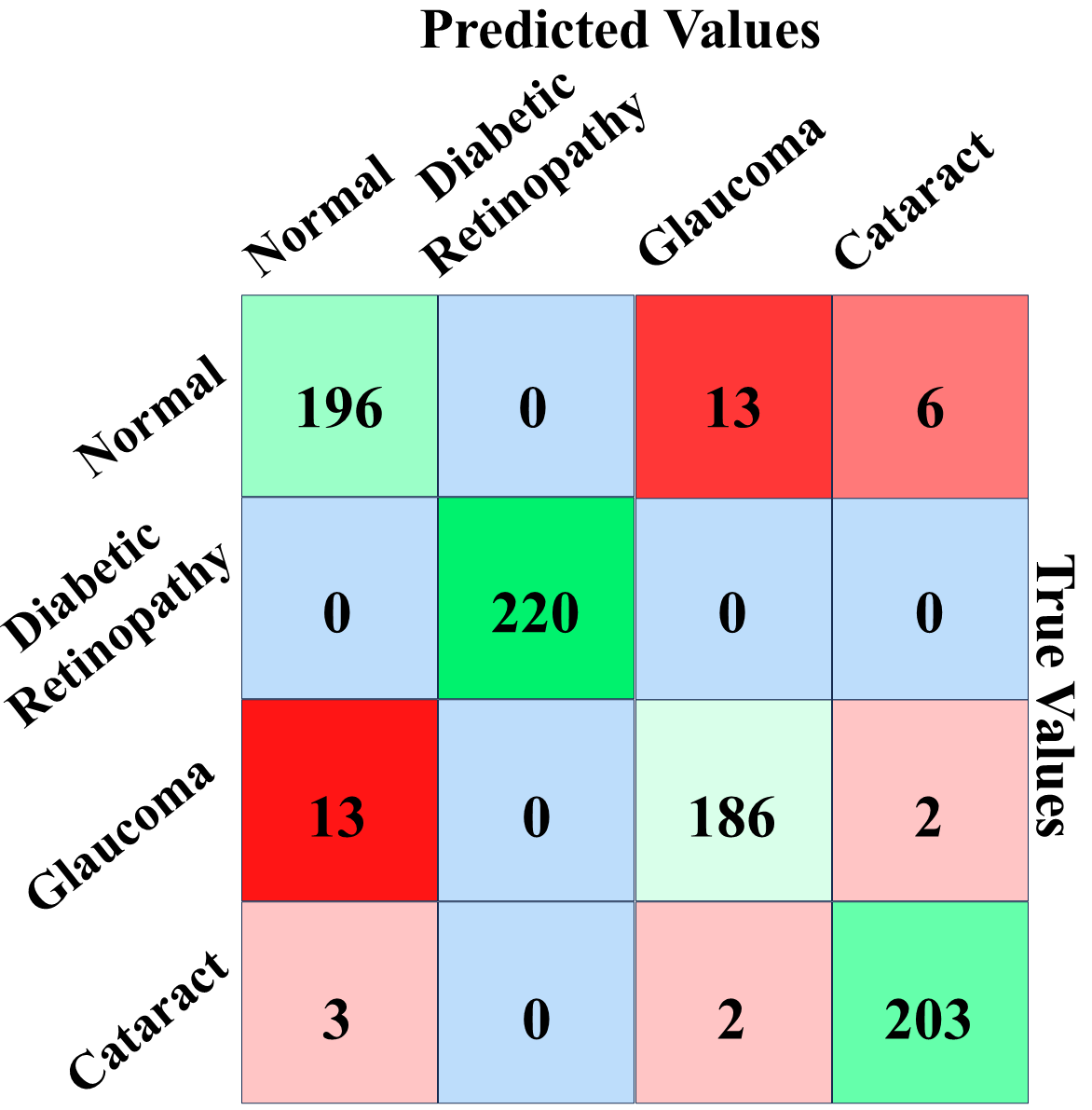}
\caption{}
\label{fig:test_cm}
\end{subfigure}

\begin{subfigure}[t]{0.45\columnwidth}
\centering
\includegraphics[
width=\linewidth,
height=0.8\linewidth,
keepaspectratio=false,
clip
]{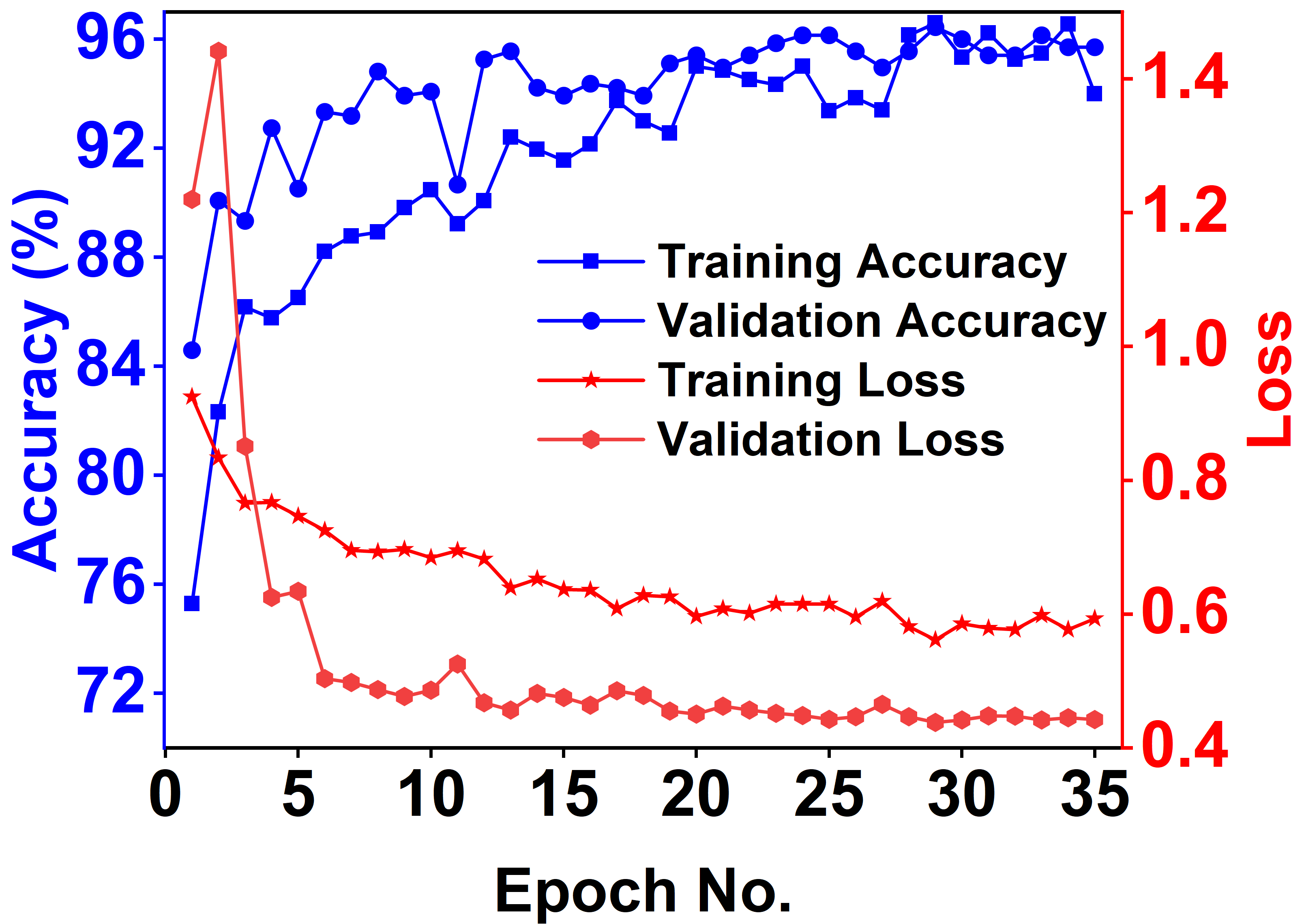}
\caption{}
\label{fig:best_fold_curve}
\end{subfigure}
\hfill
\begin{subfigure}[t]{0.45\columnwidth}
\centering
\includegraphics[
width=\linewidth,
height=0.8\linewidth,
keepaspectratio=false,
clip
]{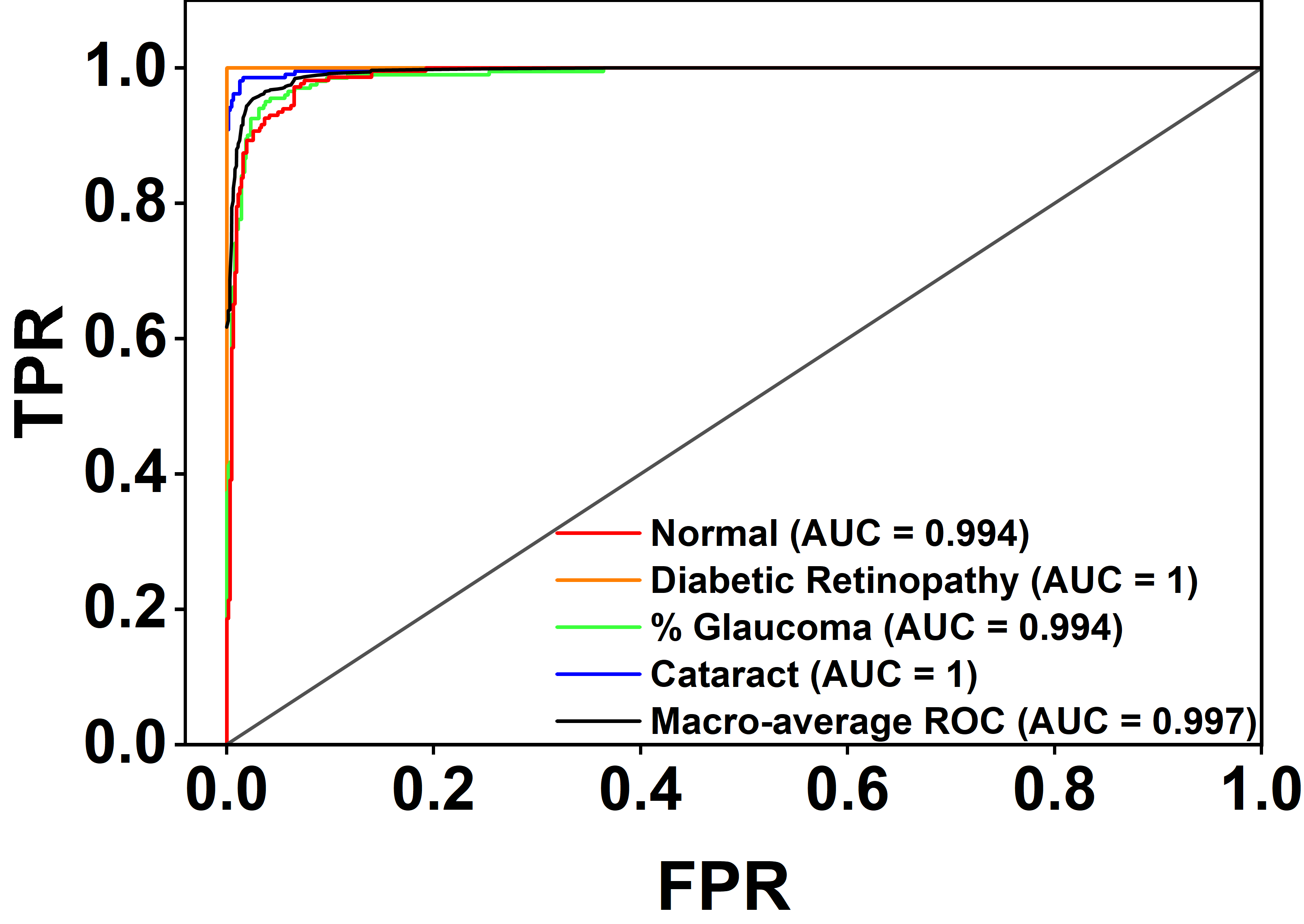}
\caption{}
\label{fig:test_curve}
\end{subfigure}

\caption{Detection performance visualization of the proposed CounterFundus framework: (a) confusion matrix for the best validation fold, (b) confusion matrix for the locked test set, (c) training/validation accuracy and loss curves for the best fold, and (d) one-vs-rest class-wise ROC curves with macro-average ROC-AUC for the locked test set.}
\label{fig:classification_results}
\vspace{-0.5em}
\end{figure}

As mentioned earlier, the EfficientNet-B5 classifier is trained using the 80\% CV pool and evaluated on the 20\% locked test set. The per-fold and test-set performances are reported in Table~\ref{tab:clf} in percentage values (mean $\pm$ std.), indicated in terms of accuracy (Acc.), precision (Prec.), recall (Rec.), F1-score (F1) and AUC. The corresponding confusion matrices, convergence curves and ROC curves are shown in Fig.~\ref{fig:classification_results}. The EfficientNet-B5 classifier is observed to converge stably across the five folds, achieving a CV accuracy of $95.2 \pm 0.69$\% and an F1-score of $95.14 \pm 0.7$\%. The limited inter-fold variation, together with the smooth training and validation trends seen in Fig.~\ref{fig:classification_results}(c), suggests that the adopted mixup and label-smoothing protocols effectively reduce overfitting. On the locked test set, the classifier achieved 95.38\% accuracy, 95.31\% macro F1-score and 99.69\% AUC, confirming robust disease-discrimination capability of CounterFundus.

To further analyze the locked test-set behaviour, class-wise performance metrics are reported in Table~\ref{tab:classwise_test}. Diabetic retinopathy is seen to achieve perfect precision, recall and F1-score of 100\%, while cataract achieved a high F1-score of 96.9\%. Normal and glaucoma obtained relatively lower F1-scores of 91.8\% and 92.54\%, respectively, mainly due to mutual class overlap. Corresponding to this particular finding, Fig.~\ref{fig:classification_results}(b) also shows 13 false positive outcomes between the normal and glaucoma categories. This behaviour is consistent with the known difficulty of distinguishing early-stage glaucoma cupping from normal optic-disc variation \cite{Bajwa_2019}.

\begin{table}[!t]
\centering
\caption{Class-wise CounterFundus Performance}
\label{tab:classwise_test}
\setlength{\tabcolsep}{0pt}
\renewcommand{\arraystretch}{1.08}
\small
\begin{tabular*}{\columnwidth}{@{\extracolsep{\fill}}cccccc@{}}
\toprule
\textbf{Class} & \textbf{Acc.} & \textbf{Prec.} & \textbf{Rec.} & \textbf{F1} & \textbf{AUC} \\
 & \textbf{(\%)} & \textbf{(\%)} & \textbf{(\%)} & \textbf{(\%)} & \textbf{(\%)} \\
\midrule
Normal     & 95.85 & 92.45 & 91.16 & 91.80 & 99.4 \\
Diab. Ret. & 100 & 100 & 100 & 100 & 100 \\
Glaucoma   & 96.45 & 92.54 & 92.54 & 92.54 & 99.4 \\
Cataract   & 98.46 & 96.21 & 97.6 & 96.9 & 100 \\
\midrule
\textbf{Macro avg.} & \textbf{97.69} & \textbf{95.3} & \textbf{95.32} & \textbf{95.31} & \textbf{99.7} \\
\bottomrule
\end{tabular*}
\vspace{-1.5em}
\end{table}

\vspace{-0.5em}
\subsection{Qualitative Analysis of The Counterfactual Generation Scheme}
In this work, counterfactual generation quality is evaluated using structural similarity index measure (SSIM) and peak signal-to-noise ratio (PSNR), computed per sample between each generated counterfactual and its corresponding original diseased image. In addition, Fréchet inception distance (FID) is calculated globally between the counterfactual set and 200 real normal fundus images using 2048-dimensional EfficientNet-B5 features. The results are reported in Table~\ref{tab:gq}. 

As shown in Table~\ref{tab:gq}, glaucoma achieves the highest SSIM of 0.82 and PSNR of 24.22 dB, consistent with localized optic-disc changes requiring limited global structural modification \cite{Bajwa_2019}. In contrast, diabetic retinopathy shows the lowest PSNR of 19.04 dB, reflecting broader vascular-region alterations associated with hemorrhages and exudates, although it maintains a moderate SSIM of 0.78. The FID of 48.56 indicates reasonable distributional alignment between generated counterfactuals and real normal images, considering the multi-disease domain gap \cite{wu2024self}. The obtained results confirm that $G_{D \to N}$ produces structurally plausible counterfactual outputs rather than arbitrary image hallucinations.

\begin{table}[!t]
\centering
\caption{Counterfactual Generation Quality}
\label{tab:gq}
\setlength{\tabcolsep}{4pt}
\renewcommand{\arraystretch}{1.08}
\small
\begin{tabular}{lccc}
\toprule
\textbf{Class} & \textbf{SSIM} & \textbf{PSNR (dB)} & \textbf{FID} \\
\midrule
Cataract      & $0.75 \pm 0.04$ & $21.64 \pm 2.42$ & \multirow{3}{*}{48.56} \\
Diab. Ret.    & $0.78 \pm 0.06$ & $19.04 \pm 3.48$ &  \\
Glaucoma      & $0.82 \pm 0.043$ & $24.22 \pm 2.69$ &  \\
\midrule
\textbf{Overall} & \textbf{0.783} & \textbf{21.63} & \textbf{48.56} \\
\bottomrule
\end{tabular}
\vspace{-0.5em}
\end{table}

\begin{table}[!t]
\centering
\caption{Classifier Confidence Shift After Translation}
\label{tab:conf}
\setlength{\tabcolsep}{4pt}
\renewcommand{\arraystretch}{1.08}
\small
\begin{tabular}{lcccc}
\toprule
\textbf{Class} & \textbf{Orig. D (\%)} & \textbf{CF D (\%)} &
\textbf{Orig. N (\%)} & \textbf{CF N (\%)} \\
\midrule
Cataract   & 89.24 & 22.17 & 3.76 & 69.64 \\
Diab. Ret. & 90.65 & 8.77  & 3.16 & 78.66 \\
Glaucoma   & 87.06 & 23.83 & 4.03 & 65.48 \\
\bottomrule
\end{tabular}
\vspace{-1.5em}
\end{table}

\vspace{-1em}
\subsection{Classifier Confidence Shift After Counterfactual Translation}
\begin{figure}[!h]
\centering
\includegraphics[
width=0.8\columnwidth,
height=0.58\columnwidth,
keepaspectratio=false,
]{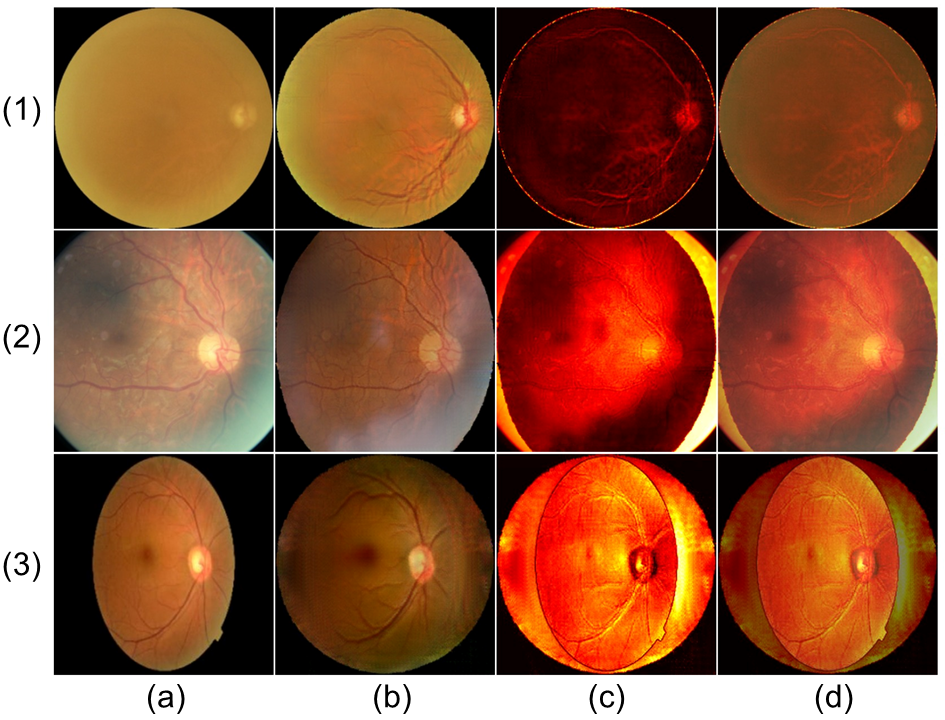}
\vspace{0.25em}
\caption{Disease-to-normal counterfactual translation results for (1) diabetic retinopathy, (2) glaucoma and (3) cataract disease classes. The Columns denote (a) original fundus image, (b) generated counterfactual, (c) corresponding difference map and (d) boundary overlays, respectively.}
\label{fig:counterfactual_new}
\vspace{-0.5em}
\end{figure}

To validate the semantic meaningfulness of disease-to-normal translation in this work, classifier softmax confidence is measured for each original diseased image and its corresponding counterfactual. Table~\ref{tab:conf} reports the average per-class confidence before and after translation, while representative disease-to-normal outputs are shown in Fig.~\ref{fig:counterfactual_new}. 

It can be noticed from Table~\ref{tab:conf} that the average disease confidence decreases by 0.67 points after translation, while the mean normal-class confidence increases by the same margin. This near-symmetric confidence shift indicates that $G_{D \to N}$ modifies disease-discriminative regions and shifts the classifier response toward the normal class, rather than producing only superficial appearance changes \cite{zhu2020unpairedimagetoimagetranslationusing}. Diabetic retinopathy shows the largest shift, with disease confidence decreasing from 90.65\% to 8.77\%, which is consistent with localized lesion patterns such as hemorrhages and exudates whose removal strongly affects the classifier decision \cite{Boreiko2022VisualExplanations}.

The visual examples illustrated in Fig.~\ref{fig:counterfactual_new} further show that the counterfactual generator largely preserves the retinal background while suppressing disease-specific regions related to the classifier decision. This qualitative observation corresponds to Table~\ref{tab:conf}, where reduced disease confidence is accompanied by a corresponding increase in normal-class confidence.

\begin{table}[!t]
\centering
\caption{CCAS Metric Validation Tests}
\label{tab:val}
\setlength{\tabcolsep}{4pt}
\renewcommand{\arraystretch}{1.08}
\small
\begin{tabular}{lccc}
\toprule
\textbf{Test} & \textbf{Variables} & \textbf{Stat.} & \textbf{$p$} \\
\midrule
T1 & IoU vs.\ Conf.\ Shift      & $\rho{=}0.05$ & 0.546 \\
T2 & Correct vs.\ Incorrect IoU & $\mu{=}0.561$  & -- \\
T3 & IoU across methods (K-W)   & $H{=}136.95$   & $<$0.0001 \\
T4 & IoU vs.\ SSIM              & $r{=}0.009$    & 0.916 \\
T5 & IoU prediction AUC         & --             & -- \\
T6 & EigenCAM vs.\ GradCAM      & 0.563/0.22    & $<$0.0001 \\
\bottomrule
\end{tabular}
\vspace{-1.25em}
\end{table}

\vspace{-0.5em}
\subsection{Validation Study based on The CCAS Metric}
In this section, six validation tests are performed to examine whether the CCAS behaves as a statistically reliable and clinically meaningful spatial alignment metric. The outcomes are summarized in Table~\ref{tab:val}. T1 evaluates the association between CCAS-IoU and classifier confidence shift, whereas T4 examines its relation with SSIM-based image fidelity. Both tests unravel weak and non-significant correlations, which is expected since CCAS-IoU measures geometric overlap between counterfactual evidence and saliency regions rather than scalar confidence variation or global image similarity. In particular, the near-zero correlation in T4, with $r(=0.009)$ and $p(=0.916)$, confirms CCAS-IoU being largely independent of image fidelity and capable of serving as a separate spatial alignment measure.

Stronger statistical evidences are observed in T3 and T6 also. In T3, the Kruskal-Wallis test reports a significant difference in IoU across saliency methods ($H=136.95$, $p<0.0001$), indicating that CCAS is sensitive to method-level localization differences. In T6, CCAS-alligned EigenCAM achieves a higher alignment score than GradCAM, yielding IoU values of 0.563 and 0.22, respectively, and the difference is statistically significant ($p<0.0001$) also. These results support CCAS being an independent spatial alignment measure for assessing the agreement between counterfactual modifications and classifier-yielded regions.

\vspace{-0.5em}
\subsection{CCAS-based Evaluation Across Disease Categories}
\label{subsec:ccas_disease_categories}
This section evaluates the spatial agreement between counterfactual difference maps and classifier-derived saliency across retinal disease categories. To avoid unstable counterfactual responses, samples having difference map standard deviations greater than 0.35 are excluded. Both the counterfactual difference map and EigenCAM saliency map are also resized to $56 \times 56$, with a circular ROI having radius of $0.35H$.

The CCAS metrics are computed using 50 samples per class and the results are reported in Table~\ref{tab:ccas}, from which Spearman correlation is seen to exceed 0.93 for all classes, indicating strong rank-order agreement between counterfactual difference maps and EigenCAM saliency. Diabetic retinopathy achieves the highest IoU@0.3 of 0.689 and $PA$ of 0.64, consistent with spatially concentrated retinal hemorrhages being localized by both the generator and classifier \cite{Boreiko2022VisualExplanations}. In contrast, glaucoma exhibits the lowest IoU@0.3 of 0.480, likely because diffuse optic-disc cupping is less sharply captured under binary thresholding set at $\tau (= 0.3)$ \cite{Bajwa_2019}.

\begin{figure}[!t]
\centering
\includegraphics[
width=0.75\columnwidth,
height=0.54\columnwidth,
keepaspectratio=false,
]{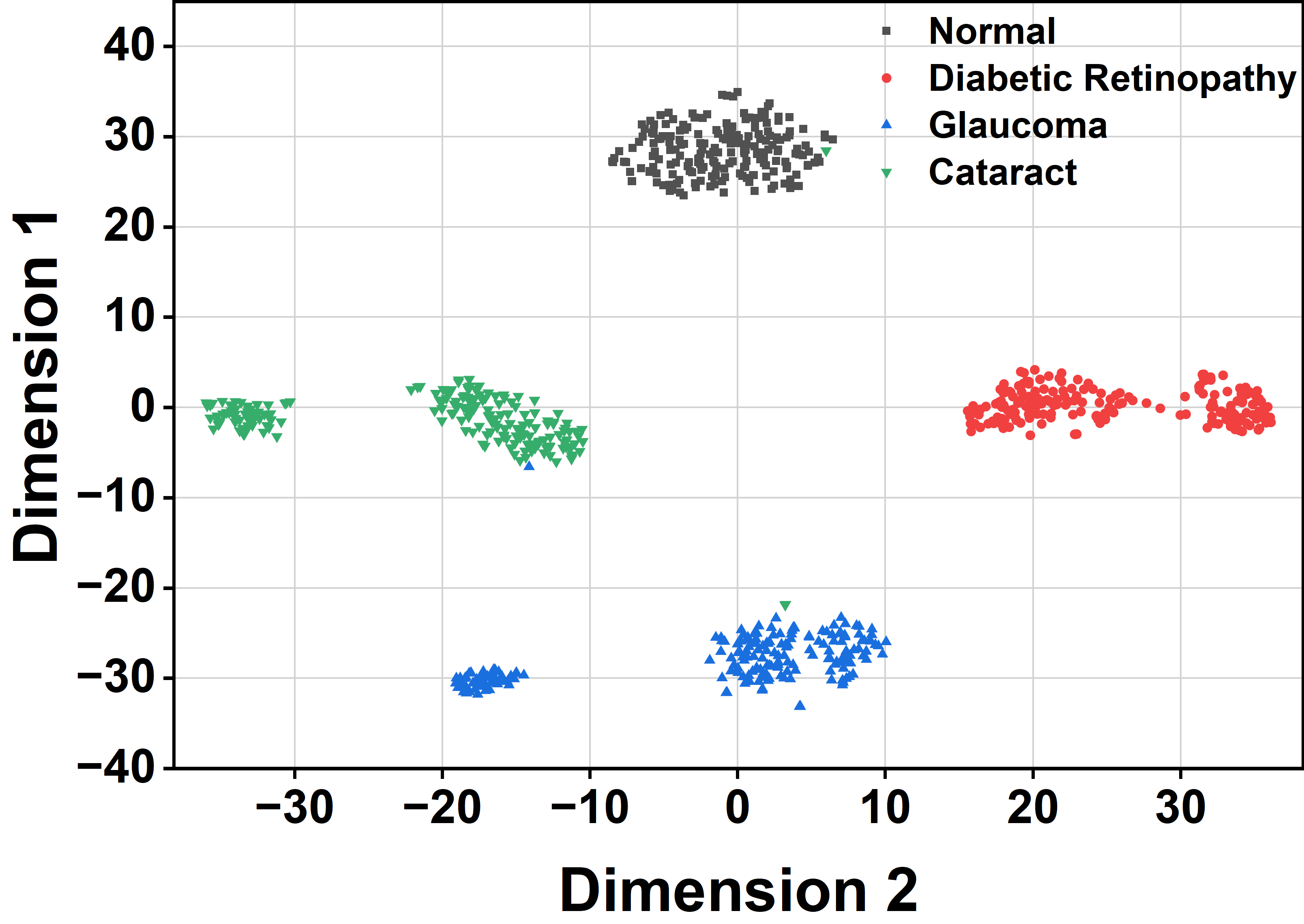}
\caption{t-SNE projection of EfficientNet-B5 features.}
\label{fig:tsne}
\vspace{-0.5em}
\end{figure}

To examine whether the classifier and generator converge toward pathology-relevant representations, Fig.~\ref{fig:tsne} shows the t-SNE projection of EfficientNet-B5 features using 200 samples per class. The well-separated clusters indicate geometrically distinct disease representations, consistent with the high CCAS across disease classes reported in Table~\ref{tab:ccas}. Together, the near-symmetric confidence shift in Table~\ref{tab:conf} and separated feature clusters in Fig.~\ref{fig:tsne} suggest that the classifier and generator independently capture overlapping pathology-relevant features \cite{zhu2020unpairedimagetoimagetranslationusing}, supporting CCAS as a reliable assessment metric.

\begin{figure}[!h]
    \centering
    \includegraphics[
width=0.8\columnwidth,
height=0.58\columnwidth,
keepaspectratio=false,
]{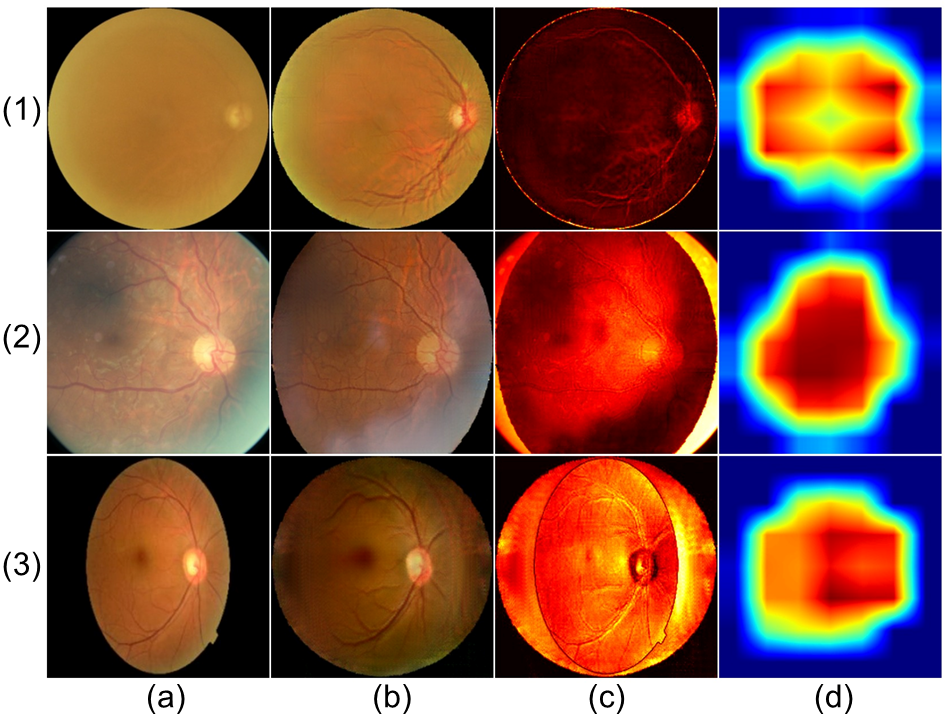}
\caption{CCAS visualization for (1) diabetic retinopathy, (2) glaucoma and (3) cataract disease classes. Columns show (a) original fundus image, (b) CycleGAN-generated normal counterfactual, (c) weighted difference map and (d) EigenCAM saliency map, respectively.}
\label{fig:ccas_viz}
\vspace{-0.5em}
\end{figure}

Further supporting this observation, Fig.~\ref{fig:ccas_viz} shows that counterfactual difference maps overlap well with classifier-discriminative regions. Moreover, EigenCAM also exhibits strong spatial agreement between the original fundus images and the counterfactuals, indicating reliable localization of pathological evidence under the proposed CCAS protocol.

\begin{table}[!t]
\centering
\caption{Per-class CCAS Results Using EigenCAM}
\label{tab:ccas}
\setlength{\tabcolsep}{4pt}
\renewcommand{\arraystretch}{1.08}
\small
\begin{tabular}{lccc}
\toprule
\textbf{Class} & \textbf{Spearman Correlation  ($\rho$)} &
\textbf{IoU@0.3} & \textbf{PA} \\
\midrule
Cataract   & 0.938 & 0.521 & 0.46 \\
Diab. Ret. & 0.959 & 0.689 & 0.64 \\
Glaucoma   & 0.952 & 0.48 & 0.54 \\
\midrule
\textbf{Overall} & \textbf{0.95} & \textbf{0.563} & \textbf{0.547} \\
\bottomrule
\end{tabular}
\vspace{-1.5em}
\end{table}

\vspace{-0.5em}
\subsection{Explainability Analysis of the Proposed CounterFundus Framework}

In this section, CCAS is evaluated as an XAI assessment measure using EigenCAM across 50 samples per class and compared against different widely used saliency methods. Among these, GradCAM and GradCAM++ are computed using class-specific targets, whereas LIME is implemented with 200 perturbation samples while retaining the top five superpixel features. The method-level aggregated CCAS results are reported in Table~\ref{tab:method}.

\begin{table}[!t]
\centering
\caption{CCAS-based Comparison of XAI Interpretability}
\label{tab:method}
\setlength{\tabcolsep}{4pt}
\renewcommand{\arraystretch}{1.08}
\small
\begin{tabular}{lccc}
\toprule
\textbf{Method} & \textbf{Spearman Correlation ($\rho$)} &
\textbf{IoU@0.3} & \textbf{PA} \\
\midrule
GradCAM    & 0.751 & 0.22 & 0.26 \\
GradCAM++  & 0.629 & 0.187 & 0.253 \\
LIME       & 0.4 & 0.2 & 0.08 \\
\textbf{EigenCAM}   & \textbf{0.95} & \textbf{0.563} & \textbf{0.547} \\
\bottomrule
\end{tabular}
\vspace{-1.5em}
\end{table}

Among the evaluated XAI methods, EigenCAM achieves the strongest performance across all CCAS metrics. LIME records the lowest pointing accuracy of 0.08, indicating that superpixel-level perturbations are less suitable for localizing continuous pathological regions in fundus images. GradCAM++ performs below GradCAM despite its higher-order gradient formulation, suggesting that first-order activation decomposition better represents the EfficientNet-B5 feature space in the present retinal disease detection task. The clear IoU@0.3 margin between EigenCAM and the gradient-based methods, i.e., 0.563 vs. 0.22, further indicates that covariance-based activation decomposition provides stronger spatial consistency with the counterfactual evidence \cite{arun2021assessinguntrustworthinesssaliencymaps}.

\begin{figure}[!t]
    \centering
    \includegraphics[
width=0.8\columnwidth,
height=0.58\columnwidth,
keepaspectratio=false,
]{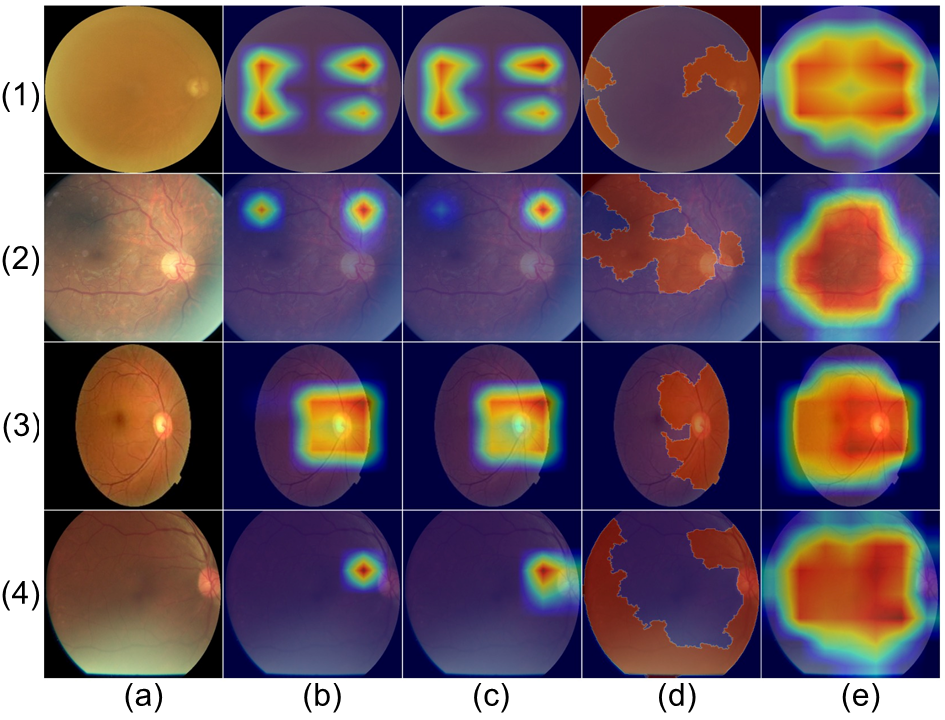}
\caption{Qualitative comparison of saliency maps across representative classes. Rows correspond to (1) normal, (2) diabetic retinopathy, (3) glaucoma and (4) cataract disease classes, while columns show (a) original fundus image, (b) Grad-CAM, (c) Grad-CAM++, (d) LIME and (e) EigenCAM overlays, respectively.}
\vspace{-0.5em}
\label{fig:baseline_grid}
\vspace{-1em}
\end{figure}

The qualitative comparison of XAI-based saliency overlays is also shown in Fig.~\ref{fig:baseline_grid}, where each row contains the original fundus image followed by GradCAM, GradCAM++, LIME and EigenCAM overlays. Compared with the other methods, EigenCAM visibly produces more coherent activation over disease-relevant retinal regions with less fragmented or displaced responses. This visual observation is again consistent with results furnished Table~\ref{tab:method}, where EigenCAM shows the strongest CCAS alignment across the evaluated metrics.

\vspace{-0.5em}
\subsection{Ablation Study}
In this work, an ablation study is also performed to examine the capability of counterfactual augmentation and CCAS-based sample filtering on locked test-set performance. For this purpose, three configurations are considered, where A1 denotes the baseline classifier trained using the original 5-fold CV protocol without counterfactual augmentation. The A2 configuration extends the training pool by adding all 150 CycleGAN-generated normal counterfactual images, with 50 samples generated from each disease class. In A3, an additional quality-control step is applied, where only 108 synthetic images satisfying the CCAS-IoU $>40\%$ criterion with EigenCAM saliency are retained.

As reported in Table~\ref{tab:ablation}, both A2 and A3 are observed to improve over the baseline A1, confirming the contribution of counterfactual augmentation towards classifier generalization. The highest accuracy, F1-score and AUC with values of 97.99\%, 97.96\%, and 99.65\%, respectively has been achieved with A2. However, A3 provides comparable performance using only 72\% of the synthetic samples, reaching 97.51\% accuracy, 97.46\% F1-score and 99.62\% AUC. The minimal accuracy difference between A2 and A3, i.e., 0.48 percentage points, indicates that CCAS-based filtering removes lower-alignment counterfactuals while preserving most of the augmentation benefit. This supports the IoU $>40\%$ threshold as a meaningful quality standard for selecting reliable counterfactual samples.

\vspace{-0.5em}
\subsection{External Validation}
External validation of the CounterFundus explainability was performed on the Retinal Fundus Multi-disease Image Dataset (RFMiD) \cite{s3g7-st65-20}, a clinically annotated benchmark with expert-verified disease labels. CCAS was computed without any parameter updates or domain adaptation, yielding Spearman $\rho = 0.48$ and IoU = 0.49. CounterFundus counterfactual explanations generalize to heterogeneous fundus distributions beyond the training domain, validating the framework's explainability transferability.
\vspace{-0.5em}
\section{Conclusion}

This work proposes CounterFundus, an explainable framework for multi-class retinal disease detection, combining an EfficientNet-B5 classifier, CycleGAN-based disease-to-normal counterfactual translation and a novel CCAS metric for spatial validation of classifier-relevant counterfactual evidence. The proposed framework demonstrates strong disease-discrimination ability while maintaining stable generalization across CV splits. The generated counterfactuals largely preserve the retinal background while reducing disease-related visual cues, including optic-disc cupping in glaucoma and opacity-related degradation in cataract. The post-translation confidence shift further supports that the generator focuses on pathology-relevant regions specifically. In addition, the proposed CCAS metric provides an effective mechanism for assessing whether the generated counterfactuals are spatially aligned with classifier-yielded saliency regions. The main findings of this work are summarized as follows:

\begin{table}[!t]
\centering
\caption{Ablation Study on Locked Test Set}
\label{tab:ablation}
\setlength{\tabcolsep}{4pt}
\renewcommand{\arraystretch}{1.08}
\small
\begin{tabular}{lccc}
\toprule
\textbf{Config} & \textbf{Acc. (\%)} & \textbf{F1 (\%)} & \textbf{AUC (\%)} \\
\midrule
A1: Classifier Only      & 95.38 & 95.31 & 99.69 \\
A2: $+$ CF Aug (150)     & 97.99 & 97.96 & 99.65 \\
A3: $+$ CCAS Filter (108) & 97.51 & 97.46 & 99.62 \\
\bottomrule
\end{tabular}
\vspace{-1.5em}
\end{table}

\begin{enumerate}[leftmargin=*, label=\arabic*), itemsep=0.2em, topsep=0.2em] 
\item The EfficientNet-B5 module achieved 95.38\% accuracy, 95.31\% macro F1-score and 99.69\% AUC on the locked test set, while the overall 5-fold CV accuracy of $95.2 \pm 0.69$\% and F1-score of $95.14 \pm 0.7$\% confirmed stable generalization across data splits.

\item Class-wise evaluation showed strong retinal disease detection performance across normal, diabetic retinopathy, glaucoma and cataract categories, with diabetic retinopathy achieving 100\% precision, recall, F1-score and AUC and cataract achieving 96.9\% F1-score and 100\% AUC.

\item The proposed CCAS metric efficiently captured the spatial agreement between counterfactual difference maps and classifier-yielded saliency regions, with EigenCAM-based reference maps achieving 0.95 Spearman correlation and 0.563 IoU@0.3 metrics, respectively.

\item The CCAS-filtered counterfactual augmentation improved classification accuracy by 2.13\% over the baseline while using 28\% fewer synthetic samples, indicating that the counterfactuals provided more reliable augmentation than unfiltered synthetic samples.

\item Among the evaluated XAI methods, EigenCAM showed the strongest CCAS alignment and outperformed GradCAM, GradCAM++ and LIME-based explanations in terms of spatial consistency with counterfactual evidence.
\end{enumerate}

Overall, CounterFundus improves both prediction performance as well as explanation reliability, providing a clinically meaningful XAI tool for retinal disease detection. Future work will include validating CCAS against different lesion-level annotations, with focusing beyond fundus imaging to different applications such as OCT, dermoscopy and chest X-ray.

\section*{References}
\vspace{-1.5em}
\phantomsection
\def\refname{References}
\bibliographystyle{IEEEtran}
\bibliography{reference}
\end{document}